\documentclass[letterpaper, 10 pt, conference]{ieeeconf}  %

\IEEEoverridecommandlockouts                              %

\usepackage{amsmath}
\usepackage{amssymb}
\usepackage{array}%
\usepackage{pifont}%
\usepackage{xcolor,colortbl}
\usepackage{xspace}
\usepackage{float}
\usepackage{comment}
\usepackage{graphicx}
\usepackage{tikz}
\usepackage{lipsum}
\usepackage[caption=false, font=footnotesize]{subfig}
\usepackage{booktabs}
\usepackage{siunitx}
\usepackage{url}
\usepackage[T1]{fontenc}

\newcommand{\cmark}{\ding{51}}%
\newcommand{\xmark}{\ding{55}}%
\definecolor{Gray}{gray}{0.9}

\usepackage{multirow}

\newcommand\copyrighttext{\footnotesize \textcopyright 2022 IEEE. Personal use of this material is permitted. Permission from IEEE must be obtained for all other uses, in any current or future media, including	reprinting/republishing this material for advertising or promotional purposes, creating new	collective works, for resale or redistribution to servers or lists, or reuse of any copyrighted component of this work in other works.
}%

\title{\LARGE \bf 
A Benchmark for Unsupervised Anomaly Detection in Multi-Agent Trajectories
}

\author{Julian Wiederer$^{1, 2}$, Julian Schmidt$^{1, 2}$, Ulrich Kressel$^{1}$, Klaus Dietmayer$^{2}$ and Vasileios Belagiannis$^{3, \dagger}$%
\thanks{$^{1}$Julian Wiederer, Julian Schmidt and Ulrich Kressel are with Mercedes-Benz Group AG, 70546 Stuttgart,~Germany. 
{\tt\small \{julian.wiederer, julian.sj.schmidt, ulrich.kressel\}@mercedes-benz.com}}%
\thanks{$^{2}$Julian Wiederer, Julian Schmidt and Klaus Dietmayer are with the Institute of Measurement, Control and Microtechnology, University Ulm, 89081 Ulm,~Germany.
{\tt\small klaus.dietmayer@uni-ulm.de}}%
\thanks{$^{3}$Vasileios Belagiannis is with the Department of Simulation and Graphics, Otto-von-Guericke-University Magdeburg, 39106 Magdeburg,~Germany.
{\tt\small vasileios.belagiannis@ovgu.de}}
\thanks{$^\dagger$Most of the work was done while VB was affiliated with Ulm University.}
}

\newcommand\copyrightnotice{%
	\begin{tikzpicture}[remember picture,overlay]
	\node[anchor=south,xshift=0pt,yshift=14pt] at (current page.south) {\fbox{\parbox{\dimexpr\textwidth-\fboxsep-\fboxrule\relax}{\copyrighttext}}};
	\end{tikzpicture}%
}

\begin{document}

\maketitle
\thispagestyle{empty}
\pagestyle{empty}

\begin{abstract}
Human intuition allows to detect abnormal driving scenarios in situations they never experienced before. Like humans detect those abnormal situations and take countermeasures to prevent collisions, self-driving cars need anomaly detection mechanisms. However, the literature lacks a standard benchmark for the comparison of anomaly detection algorithms. We fill the gap and propose the R-U-MAAD benchmark for unsupervised anomaly detection in multi-agent trajectories. The goal is to learn a representation of the normal driving from the training sequences without labels, and afterwards detect anomalies. We use the Argoverse Motion Forecasting dataset for the training and propose a test dataset of 160 sequences with human-annotated anomalies in urban environments. To this end we combine a replay of real-world trajectories and scene-dependent abnormal driving in the simulation. In our experiments we compare 11 baselines including linear models, deep auto-encoders and one-class classification models using standard anomaly detection metrics. The deep reconstruction and end-to-end one-class methods show promising results. The benchmark and the baseline models will be publicly available\footnote{Project page: \url{https://github.com/againerju/r_u_maad}}.
\end{abstract}

\section{Introduction}

\copyrightnotice

Human drivers detect abnormal driving situations and immediately reduce velocity to prevent collisions. A self-driving car must contain a similar anomaly detection mechanism to supplement the perception and forecasting modules~\cite{liang2020learning, schmidt2022cratpred, schreiber2020motion, strohbeck2020Multiple, wiederer2020traffic}. Recently, anomaly detection in traffic has gained interest~\cite{matousek2019detecting, oikawa2020fast, wang2016statistical, pmlr-v164-wiederer22a}, however the literature lacks an annotated real-world benchmark with diverse anomalies for comparison of the algorithms. 

In this work, we propose a benchmark for unsupervised anomaly detection in traffic scenes. 
It extends the Argoverse Motion Forecasting benchmark~\cite{chang2019argoverse} to anomaly detection. The goal is to learn a representation of the normal behaviour only from real-world trajectory data without anomaly labels and to detect anomalies in our test set with manually annotated normal and abnormal driving behaviour. We compare linear models, deep auto-encoders and one-class classification models as baselines on our benchmark. 

\begin{figure}[t]
    \centering
    \includegraphics[width=\linewidth]{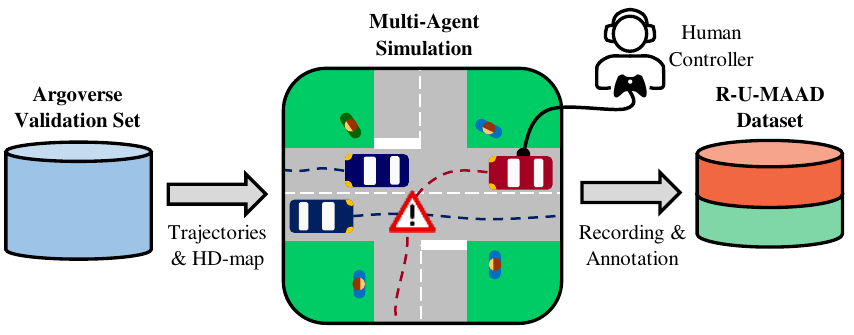}
    \caption{Our pipeline for abnormal trajectory generation. For a sequence from the Argoverse validation set, we visualise the scene from the HD-map in our multi-agent simulation and hijack one vehicle (red), the target vehicle, for animation. A human controller drives abnormal scenarios, while the remaining agents (blue) are replayed from the trajectory as recorded in the Argoverse dataset. We record the trajectory of the hijacked vehicle and conduct manual annotation of anomaly intervals. In addition, we drive normal scenarios to complete our dataset. This results in a diverse dataset for multi-agent anomaly detection in realistic urban settings, we call R-U-MAAD dataset, with normal (green) and abnormal (orange) samples.}
    \label{fig:data_generation}
\end{figure}

There is a rich literature on the prediction of future trajectories, mainly focused on real-world trajectories and HD-maps~\cite{nuscenes, chang2019argoverse, interactiondataset}. %
However, the majority of the driving scenarios show normal behaviour and they do not consider abnormal situations. Abnormal driving does not follow the normal driving rules and therefore the trajectory predictions may be inaccurate~\cite{makansi2022you}. These situations can be identified using anomaly detection. 

We present a benchmark with abnormal situations for anomaly detection. Similar to standard protocols in trajectory prediction~\cite{gao2020vectornet, liang2020learning}, we describe scenes with agent trajectories and HD-maps. Since anomalies are rare in the real-world and staging of abnormal scenarios is dangerous for the actors, simulation is employed to create driving anomalies~\cite{sabour2021deepflow, shah2021asimulation, ucar2021differential, pmlr-v164-wiederer22a}.
Instead of relying on simulation only, we propose to leverage as much real-world data as possible to do anomaly detection. In the training, we only use real-world data to learn from the normal behaviour, without any simulated data. In the testing, we detect anomalies in our test sequences. To create a test scenario, we take a real-world sequence and only animate a single car for the anomalies in the simulation. In addition to individual driving anomalies, we focus on anomalies in the interaction between agents, e.g.~a ghost driver, in complex urban settings with many agents on the street. To create the benchmark, we use the well-known Argoverse dataset~\cite{chang2019argoverse}, as it provides multi-agent trajectories in diverse traffic scenes and detailed scene descriptions with HD-maps. Figure~\ref{fig:data_generation} illustrates the data generation process.
Abnormal situations can happen at any driving time, e.g. a car can break rapidly without a warning or an inattentive truck driver can enter the wrong lane and drive against traffic. To evaluate anomaly detection continuously, we provide frame-wise human annotations for all anomalies in our benchmark.

To summarise our contributions, we present a benchmark for unsupervised multi-agent anomaly detection including detailed training and evaluation protocols. For training, we use real-world trajectories only. For testing, we create a hybrid test set. Each test sequence contains the animated trajectory of the target vehicle and the real-world trajectories of the remaining actors. 
In addition, we evaluate a set of algorithms including linear models, deep auto-encoders and one-class classification models for anomaly detection following the proposed protocol. The algorithms learn a representation of the normal trajectories from the training set and identify anomalies as outliers during testing. %

\section{Related Work}
Anomaly Detection (AD) is a long-standing problem with many applications in diverse fields. It is used to localise tumours in medical imaging~\cite{tian2021constrained}, to identify fake news in social networks~\cite{li2021unsupervised} and to uncover payment card frauds~\cite{branco2020interleaved}. Below, we discuss the related methods and datasets and focus on anomaly detection in the field of automated driving.

\textbf{Anomaly Detection in Automated Driving.}
Recently, anomaly detection has been applied to automotive related tasks, for example to detect brake light transitions of leading vehicles~\cite{weis2017anomaly}, failures in the brake system of a bus~\cite{cheifetz2011apattern} or traffic incidents~\cite{zhu2018adeep}. For self-driving cars the detection of anomalies in traffic is of particular interest. Every missed abnormal situation, like a driver who is distracted by writing a text message on the phone, can have fatal consequences to all participants in the scene. Often machine learning methods are explored to classify abnormal driving behaviour using explicit behaviour annotations during training on simulated datasets~\cite{oikawa2020fast, wang2016statistical}. However, by definition, anomalies are rare in the real-world and the annotation of anomalies in the entire training set is an exhaustive task. Therefore, we focus on the task of unsupervised anomaly detection. Currently, unsupervised representation learning techniques like deep auto-encoder networks determine the anomaly detection domain~\cite{eskin2002geometric, pmlr-v80-ruff18a, zong2018deep}. For the detection of driving anomalies auto-encoders with a single trajectory as input and hand-crafted features are employed~\cite{matousek2019detecting}.
Driving anomalies often occur in interaction between vehicles, for example a ghost driver on a collision course with the oncoming traffic. To this end, spatio-temporal graph auto-encoders, considering the trajectories of all agents in the scene, have been developed by Wiederer~\textit{et al.}~\cite{pmlr-v164-wiederer22a}. Based on simulation, they create a highway dataset with two agents in the scene for evaluation. Similar, our baselines model the interaction between agents, but in complex urban environments with many agents on the street.

\textbf{Anomaly Detection Datasets in Automated Driving.}
In trajectory forecasting, state-of-the-art datasets contain large sets of real-world traffic scenes with high-quality tracks of all agents and HD-maps describing the static context~\cite{nuscenes, chang2019argoverse, ettinger2021large, houston2020lyft, ma2019trafficpredict, yu2020bdd100k, interactiondataset}. However, they do not consider the critical cases of anomalies. Although some of them include critical situations and near-collision scenarios, the number is small and they do not contain behaviour annotations what limits the anomaly detection. Since anomalies are rare and staging is prohibitive due to danger for the actors, in many cases simulation is leveraged to create abnormal driving data on the cost of realism, for example by tuning the parameters of traffic and vehicle models~\cite{matousek2019detecting, sabour2021deepflow, ucar2021differential}. A more realistic approach was proposed by Shah~\textit{et al.}~\cite{shah2021asimulation}, they record anomalous scenarios with the scene description language Scenic and the CARLA simulator. Wiederer~\textit{et al.}~\cite{pmlr-v164-wiederer22a} employ an OpenAI Gym simulation to generate driving anomalies in an interactive highway scenario, where all vehicles are controlled by human players. Instead of relying on simulation only, in this work, we propose to use mostly real-world trajectory data and reduce simulation to a minimum. More specific, we use the real trajectories of a trajectory prediction dataset as unsupervised training data. Next, for the creation of the test set we replay real trajectories in our simulation and animate only one agent to drive anomalies by the control of a human (see Fig.~\ref{fig:data_generation}). Using this hybrid process, the source of all training and test data is the real-world, except for the abnormal trajectories, which are animated by a human player. In addition we provide dense human annotations of all anomalies and a detailed evaluation protocol.

\begin{figure*}[t]
\centering
    \subfloat[Normal scenario: Left turn.]{
        \includegraphics[width=0.3\textwidth]{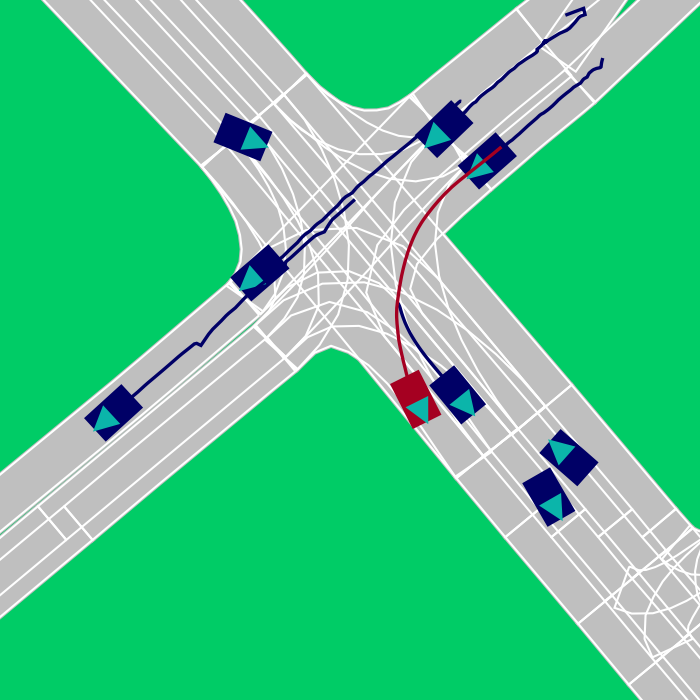}
    }
    \hfill
    \subfloat[Abnormal scenario: Ghost driving.]{
        \includegraphics[width=0.3\textwidth]{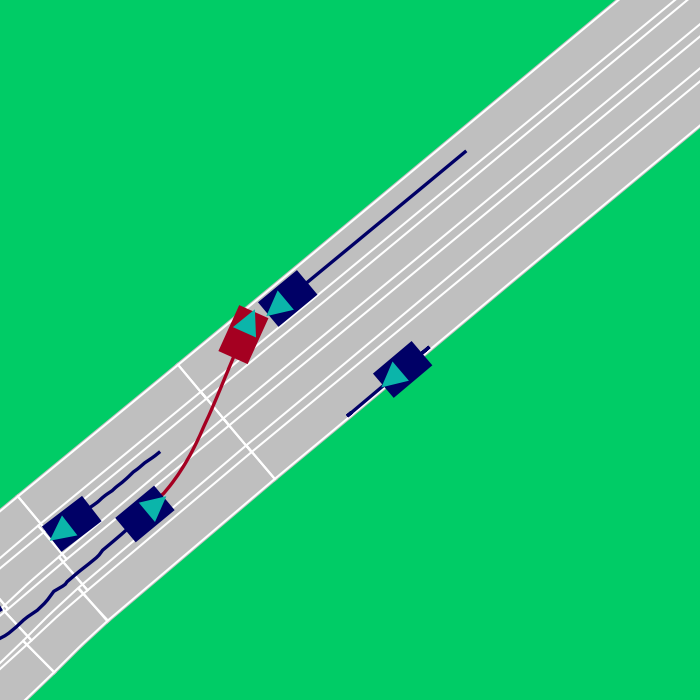}
    }
    \hfill
    \subfloat[Abnormal scenario: Aggressive shearing.]{
        \includegraphics[width=0.3\textwidth]{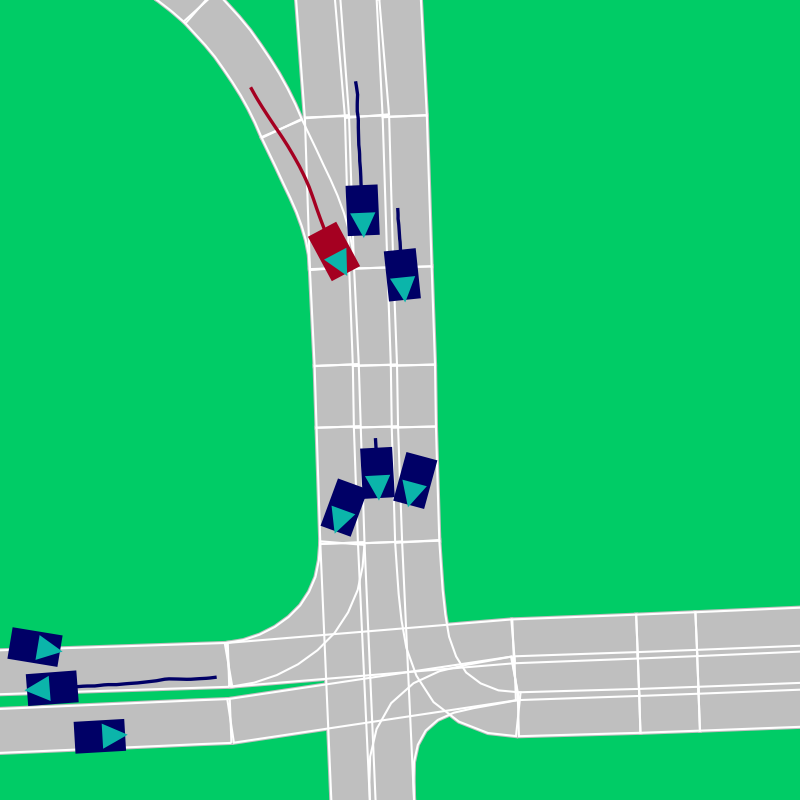}
        \label{fig:aggr_shearing}
    }
    \caption{Example scenarios from the proposed anomaly detection test set. It shows the road topology with the lane polygons in grey, the contours of the lane polygons in white and all agents in the scene with their path history. The target agent is shown in red, all remaining agents in blue. A turquoise triangle indicates the current driving direction estimated from the last two time steps.}
    \label{fig:data_samples}
\end{figure*}

\section{Multi-Agent Anomaly Detection Benchmark}

We introduce a benchmark in \textbf{R}eal-world \textbf{U}rban settings for \textbf{M}ulti-\textbf{A}gent \textbf{A}nomaly \textbf{D}etection, namely the R-U-MAAD benchmark. 
The remaining sections present the formation, the annotation and the properties of the R-U-MAAD benchmark.

\subsection{Benchmark Formation} 
We consider multi-agent anomaly detection as an unsupervised learning problem with anomaly labels only during testing. As a result, our benchmark is composed of a training and validation set without labels and a test set with anomaly labels for evaluation. We extend the Argoverse Motion Forecasting dataset~\cite{chang2019argoverse} to the task of unsupervised anomaly detection. The Argoverse Motion Forecasting dataset provides a large number of vehicle trajectories in diverse urban settings. In the original setting, the task is to forecast the trajectory of the \textit{target vehicle}, in the next $3$ seconds given $2$ seconds of history. For our benchmark, we use the training and validation sequences as provided in the Argoverse dataset during training. For testing, we create scenes with diverse labelled anomalies.

\textbf{Test Set for Anomaly Detection.} Based on a recording script for the driving scenarios, we record both normal and abnormal sequences in diverse urban settings. We create abnormal behaviour under the assumption that the abnormal driving happens before the neighbouring vehicles can react to it. This is often the case in real driving where anomalies occur suddenly with limited reaction time. Therefore, we make use of driving scenes recorded in real traffic and add driving anomalies in the simulation. To this end, we replay all agents of a scene in the OpenAI Gym simulation~\cite{brockman2016openai} and remotely control one actor by a human player. We use a subset of the Argoverse validation set as pre-recorded scenes to create our anomaly detection test set. The data generation process is visualised in Figure \ref{fig:data_generation}.
A powertrain model translates the discrete keyboard commands \textit{\{gas, break, steer left, steer right\}} into forces and a kinematic car model transforms the forces into smooth trajectories. The parameters of the car model are tuned to create motion dynamics, e.g. velocity and acceleration, similar to the original Argoverse trajectories. The initial conditions of the animated actor are set to the position and velocity of the original Argoverse~\textit{AGENT} estimated from the first two time steps. We record the 2D birds-eye view centre coordinates of our animated vehicle at \SI{10}{Hz} temporally synchronised with the other agents. We follow the Argoverse protocol and select the \textit{target vehicle} for animation. Figure~\ref{fig:data_samples} illustrates samples of our dataset.

\subsection{Normal and Abnormal Driving Annotation}
We define abnormal driving as the set of all uncommon and rare scenarios and provide a test set with frame-wise annotations of diverse driving anomalies. It includes 
driving against social rules (e.g. thwarting a following vehicle), violating regulatory driving rules (e.g. entering the wrong lane after a turn), aggressive behaviour (e.g. driving close to the vehicle ahead in order to signal the intention of an overtake) or inattentive behaviour (e.g. leaving the road because of drowsiness).
In total, this results in $13$ anomaly classes\footnote{The abnormal classes are: \textit{ghost driver, leave road, thwarting, cancel turn, last minute turn, enter wrong lane, staggering, pushing away, swerving left and right, aggressive shearing left and right and tailgating.}}. Besides individual driving anomalies, like staggering, the test set contains many anomalies which are abnormal in the context of the neighbouring agents (actor-interactive), like tailgating, abnormal in the context of the map information (map-interactive), like leaving the road, or abnormal with respect to both the static and the dynamic context, like a ghost driver. Table \ref{tab:anomaly_classes} shows the abnormal classes and the categorisation into actor-interactive and map-interactive. Two manoeuvres, \textit{swerving left} and \textit{right}, can be either interactive or non-interactive, depending on the reason for the swerving manoeuvre, e.g. another vehicle or inattentiveness. Actor- and map-interactive anomalies favour the development of interaction-aware anomaly detection.

Similar to~\cite{pmlr-v164-wiederer22a} we annotate the normal driving as well. The normal manoeuvres are described by $9$ classes\footnote{The normal classes are: \textit{straight, turn left and right, following, side-by-side, lane change left and right, brake and accelerate.}}. The annotation of both the positive and negative classes supports detailed failure analysis in cases where an abnormal class is confused with a normal class. 
For example, an algorithm ignoring the neighbouring agents is likely to confuse an aggressive shearing manoeuvre with a normal lane change. Aggressive shearing happens in close distance to the oncoming traffic (see Figure~\ref{fig:aggr_shearing}), whereas a normal lane change keeps sufficient safety distance. Our label definition allows to uncover this confusion. 
Besides normal or abnormal annotations, we label ignore regions in cases where a categorisation of pure normal or pure abnormal is not possible, e.g. in transition phases. Ignore regions are not considered during evaluation, such that errors in ignored time steps do not contribute to the result.

Our annotation process is completely manual. Human annotators watch the video of a scene from the birds-eye perspective as shown in Figure~\ref{fig:data_samples} and annotate intervals with the defined classes. All annotators are provided with detailed label instructions to ensure high label quality. We use ELAN~\cite{brugman2004annotating} as the software for annotating time series.

\subsection{Dataset Properties}

The R-U-MAAD dataset is a composed of the $205,942$ train and $39,472$ validation sequences from the Argoverse dataset and our test sequences. We record $160$ test sequences with $80$ normal and $80$ abnormal scenes. The sequences range from $1.7$ to $10.1$ seconds in length\footnote{Note, some sequences in the Argoverse dataset are longer than $5$ seconds.}. In total $7,545$ time steps are annotated with $1,412$ abnormal, $5,695$ normal and $438$ ignore labels. Table~\ref{tab:anomaly_classes} shows the distribution of the abnormal categories. 
In particular the manoeuvres \textit{ghost driver}, \textit{leave road}, \textit{thwarting} and \textit{cancel turn} occur more often as they are usually longer compared to the underrepresented categories like \textit{swerving} or \textit{aggressive shearing} manoeuvre. In total each class consists of a sufficient amount of annotations to evaluate anomaly detection algorithms.

\begin{table}[t]
\caption{The distribution of the abnormal sub-classes. We distinguish between actor-interactive and map-interactive anomalies.}
\centering
\begin{tabular}{lccc}
\toprule
\textbf{Class}          & \textbf{\begin{tabular}[c]{@{}c@{}}Actor- \\ Interactive\end{tabular}} & \textbf{\begin{tabular}[c]{@{}c@{}}Map- \\ Interactive\end{tabular}} & \textbf{\begin{tabular}[c]{@{}c@{}}\# of \\ Timesteps\end{tabular}}\\
\midrule
\rowcolor{Gray}
ghost driver            & \cmark                          & \cmark                        & 202                      \\
leave road              & \xmark                          & \cmark                        & 186                  \\
\rowcolor{Gray}
thwarting               & \cmark                          & \xmark                        & 179                      \\
cancel turn             & \xmark                          & \cmark                        & 156                      \\
\rowcolor{Gray}
last minute turn        & \xmark                          & \cmark                        & 114                      \\
enter wrong lane        & \cmark                          & \cmark                        & 101                      \\
\rowcolor{Gray}
staggering              & \xmark                          & \xmark                        & 92                      \\
pushing away            & \cmark                          & \xmark                        & 84                      \\
\rowcolor{Gray}
swerving (l)            & (\cmark)                        & \xmark                        & 77                      \\
swerving (r)            & (\cmark)                        & \xmark                        & 26                      \\
\rowcolor{Gray}
tailgating              & \cmark                          & \xmark                        & 69                      \\
aggr. shearing (l) & \cmark                          & \xmark                        & 62                       \\
\rowcolor{Gray}
aggr. shearing (r) & \cmark                          & \xmark                        & 64                      \\
\bottomrule
\end{tabular}
\label{tab:anomaly_classes}
\end{table}

\section{Problem Formulation}
The task of unsupervised anomaly detection in agent trajectories is defined as follows:
Agent $i$ has the state $\mathbf{s}_i^{t} = (x_i^{t}, y_i^{t})$, consisting of its $x$ and $y$ coordinates at timestep $t$ in a birds-eye perspective.
The trajectory of an actor $i$ is defined as $\mathbf{S}_i = \{ \mathbf{s}_i^{t} | t=-T+1, \dots ,0 \}$, with $T$ being the number of observed timesteps.
Available input data for the task of anomaly detection in multi-agent trajectories are the past trajectories $\mathbf{S} = \{ \mathbf{S}_i | i = 1, \dots, N \}$ of $N$ agents and additional context information $\mathcal{I}$.
Context information $\mathcal{I}$ is most commonly provided by an HD-map.
The goal of anomaly detection is to use this given input data in order to predict an anomaly score $\alpha_i$ of agent $i$ behaving abnormally. We present different models for anomaly detection in the following.

\section{Unsupervised Anomaly Detection Baselines}
\label{sec:models}
\begin{figure*}[t]
    \centering
    \includegraphics[width=\linewidth]{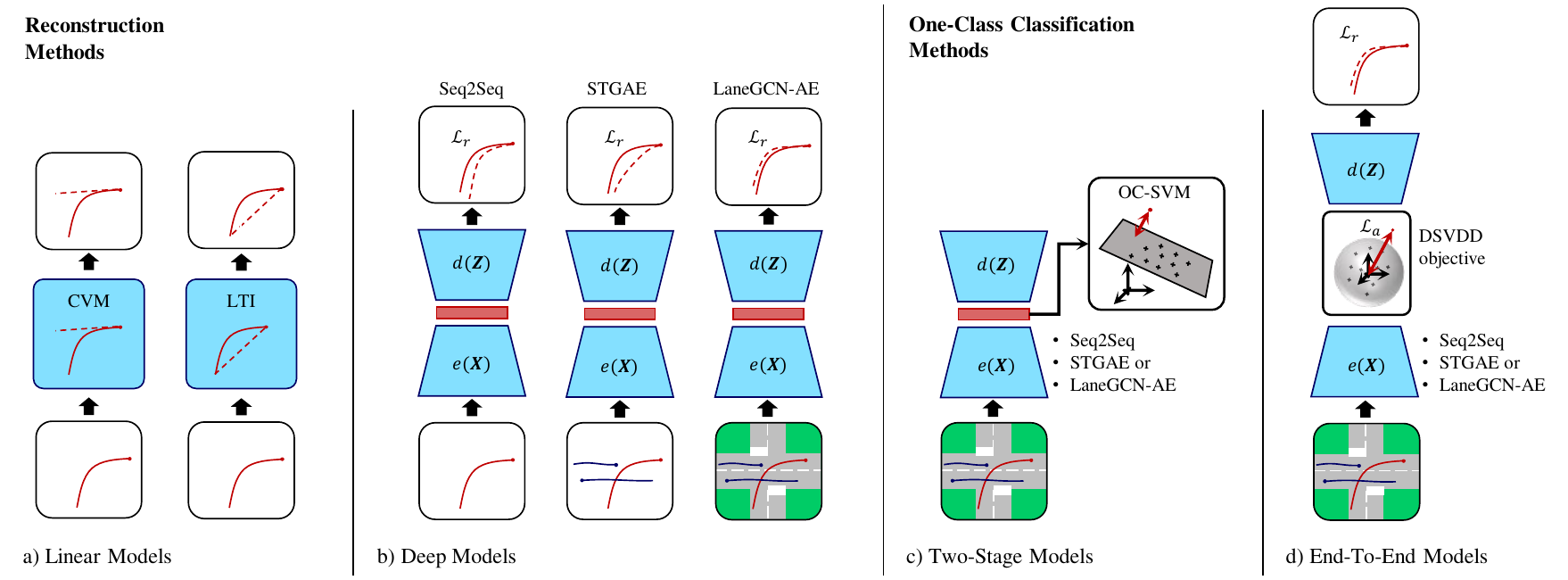}
    \caption{Our proposed baselines for anomaly detection in multi-agent trajectories. We distinguish between reconstruction and one-class classification methods. a) The linear models use the error between a linear approximation and the input trajectory as anomaly score. b) Other the deep reconstruction models, which use deep auto-encoders with a single input trajectory, multiple input trajectories or all trajectories in the scene plus the HD-map, to reconstruct the path of the target agent. c) The two-stage one-class classification methods output an anomaly score as distance from a hyperplane estimated by an OC-SVM. d) The end-to-end models optimise the DSVDD objective $\mathcal{L}_a$ jointly with the reconstruction loss $\mathcal{L}_r$ and use the distance from a hypersphere as anomaly score.}
    \label{fig:baselines}
\end{figure*}

As baselines for our anomaly detection benchmark, we evaluate $11$ different anomaly detection models including reconstruction and one-class classification methods. For both types, temporal trajectory modelling captures the temporal dependencies in the anomalies. Figure~\ref{fig:baselines} illustrates the proposed baselines.

\subsection{Reconstruction Methods}
Reconstruction methods rely on the error between the input trajectory $\mathbf{S}_i$ and an approximation of the input trajectory $\hat{\mathbf{S}}_i$ as a score for anomalies $\alpha = \lVert \mathbf{S}_i-\hat{\mathbf{S}}_i\rVert$. We compare two concepts to approximate the input trajectory, linear approximation and deep neural network auto-encoders. The linear models include:
\begin{itemize}
    \item \textbf{CVM}: The constant velocity model from~\cite{scholler2020constant}. Estimate the velocity $v_i$ from the first two time steps $t=\{-T+1, -T+2\}$ and extrapolate for $T-1$ time steps starting at the first state $\mathbf{s}_i^{-T+1}$ using $v_i$ as constant velocity.  
    \item \textbf{LTI}: The linear temporal interpolation model (\textit{LTI}) as defined in~\cite{pmlr-v164-wiederer22a}. Approximate the trajectory as equidistant samples on a straight line between the location at the first time step $\mathbf{s}_i^{-T+1}$ and the current time step $\mathbf{s}_i^{0}$. 
\end{itemize}
For the deep neural networks, we develop three auto-encoder networks, one interaction-free and two interaction-aware models. Given the input $\mathbf{X}$ the auto-encoder computes a latent representation $\mathbf{Z} = e(\mathbf{X})$ with the encoder. Based on the latent representation $\mathbf{Z} \in \mathbb{R}^F$ with feature dimension $F$ the decoder reconstructs the input $\hat{\mathbf{X}} = d(\mathbf{Z})$. Both encoder and decoder are designed as neural networks and trained end-to-end with a similarity-based loss, e.g. the MSE loss. The deep auto-encoders are defined as follows:
\begin{itemize}
    \item \textbf{Seq2Seq}: The interaction-free model takes a single agent trajectory, the trajectory of the target agent as input, $\mathbf{X}=\mathbf{S}_{i=1}$. We use the motion forecasting baseline from Argoverse, which is composed of the LSTM encoder and the LSTM decoder, and adapt it for trajectory reconstruction~\cite{chang2019argoverse}.
    \item \textbf{STGAE}:  Our first interaction-aware model is a variant of the spatio-temporal graph auto-encoder, which is used to detect anomalies in highway scenarios~\cite{pmlr-v164-wiederer22a}. Given the trajectories of all agents as input, $\mathbf{X}=\mathbf{S}$, it aggregates neighbouring agent features with the graph convolutional operator and a linear residual connection~\cite{pmlr-v164-wiederer22a}. We use the \textit{Seq2Seq} model as described above for the temporal encoding and decoding.
    \item \textbf{LaneGCN-AE}: Our second interaction-aware model takes all trajectories $\mathbf{S}$ and the context $\mathcal{I}$, given by the HD-map, as input, $\mathbf{X}=\{\mathbf{S},\mathcal{I}\}$. We adapt the motion forecasting model LaneGCN for trajectory reconstruction~\cite{liang2020learning}. Therefore we denote the original FusionNet, a graph neural network for actor and map encoding, as encoder and reduce the multiple prediction headers to a single reconstruction header.
\end{itemize}

Given the additional information on neighbouring actors or the scene, both interaction-aware models are capable to learn from interaction. The auto-encoders are trained with the MSE loss $\mathcal{L}_r = \frac{1}{T} \sum_{t=0}^{T-1}  \lVert\mathbf{s}_i^t - \hat{\mathbf{s}}_i^t \rVert$ on the reconstruction of the target agent states $\hat{\mathbf{s}}_i^t$ with $i=1$.

\subsection{One-Class Classification Methods}
One-class classification methods follow a different principle to detect anomalies. The goal is to learn a model which describes the normal data accurately. Samples which do not follow this description are scored as anomalies given by a distance measure. For the one-class classification methods we develop two-stage models and models with end-to-end training. For the two-stage models we train the one-class SVM (OC-SVM) on the latent representation $\mathbf{Z}$ of the entire training set~\cite{scholkopf1999support}. Note, we first train the auto-encoder and second train the one-class method. The OC-SVM finds a hyperplane, which represents most of the input data samples and keeps a maximum distance to the origin. Anomalies are scored by the distance to this hyperplane. We propose two-stage models for the interaction-free and interaction-aware encoders:
\begin{itemize}
    \item \textbf{Seq2Seq+OC-SVM}: The OC-SVM is trained on the interaction-free features of the target agent computed by the Seq2Seq encoder.
    \item \textbf{STGAE+OC-SVM}: Similar to the OC-SVM variant in~\cite{pmlr-v164-wiederer22a}, we train the OC-SVM on the neighbour-aware features of the target agent as provided by the STGAE encoder.
    \item \textbf{LaneGCN-AE+OC-SVM}: With the OC-SVM using the neighbour- and map-aware target agent encodings of the LaneGCN-AE encoder.
\end{itemize}

The second type of one-class methods are auto-encoder networks jointly trained with the reconstruction loss $\mathcal{L}_r$ and an anomaly detection based objective $\mathcal{L}_a$. We use the one-class deep support vector data description objective~(\textit{DSVDD}) proposed by~\cite{pmlr-v80-ruff18a} as the anomaly detection based objective. The DSVDD objective is applied on the latent representation $\mathbf{Z} \in \mathbb{R}^F$ of the auto-encoder penalising the distance of the representation from a centre $\mathbf{c} \in \mathbb{R}^F$. It tries to optimise for a hypersphere with minimum volume encapsulating all latent representations of the training data. During testing, we compute the anomaly score $\alpha = \lVert \mathbf{Z}_i - \mathbf{c} \rVert$ as the Euclidean distance of an agent feature vector $\mathbf{Z}_i$ from the centre. We define the following models:
\begin{itemize}
    \item \textbf{Seq2Seq+DSVDD}: Applying the DSVDD loss on the latent features of the Seq2Seq.
    \item \textbf{STGAE+DSVDD}: Applying the DSVDD objective on the latent interaction-aware features of the STGAE.
    \item \textbf{LaneGCN+DSVDD}: Applying the DSVDD loss on the map- and actor-dependent features of the LaneGCN-AE.
\end{itemize}

All networks are trained end-to-end minimising the un-weighted sum of both losses $\mathcal{L} = \mathcal{L}_r + \mathcal{L}_a$.

\section{Experimental Setup}

This section describes the training and evaluation protocol for unsupervised anomaly detection and the implementation details of our baseline models. 

\subsection{Training and Evaluation Protocol}
We train and validate the algorithms on the original training and validation sets of the Argoverse dataset in an unsupervised manner. For the testing, our simulated test set is used. We use a history of \SI{1.6}{s}, which is in similar range to those of publicly available trajectory prediction datasets, such as Argoverse~(\SI{2}{s})~\cite{chang2019argoverse} and WAYMO~(\SI{1}{s})~\cite{ettinger2021large}. We randomly clip a segment of $1.6$ seconds, $T=16$ frames, from each sequence in the training and validation set. For the testing, we predict an anomaly score in each time step using a sliding window approach of length $T=16$ and stride $s=1$. We do not predict an anomaly score for the first $15$ time steps of a sequence to avoid padding. For the reconstruction methods, the anomaly score is equal to the reconstruction loss, for the one-class methods, the anomaly score is equal to the distance of the feature vector from the hyperplane or from the hypersphere centre for the OC-SVM and the DSVDD variants, respectively.

To evaluate the algorithms, we use the standard anomaly detection metrics~\cite{corbiere2019addressing, zaheer2020old, pmlr-v164-wiederer22a}:
AUPR-Abnormal, AUPR-Normal, AUROC and FPR-95\%-TPR. In classification problems with imbalanced datasets AUPR and AUROC tell different stories. While the AUROC metric tries to cover both classes in one score, AUPR focuses on the positive minority class. Therefore, we define the Area Under the Precision-Recall (AUPR) curve as our main metric, since it is able to adjust for class imbalances, which is the case in anomaly detection. Our test set contains around \SI{80}{\%} normal and \SI{20}{\%} abnormal time steps. In the case of anomaly detection in automated driving, leaving the abnormal situation undetected can lead to critical situations and is more severe compared to false alarms. However, for a complete evaluation we show both AUPR scores, the AUPR-Abnormal, where we treat the abnormal class as positive, and the AUPR-Normal, where we treat the normal class as positive. The best anomaly detector has the highest AUPR-Abnormal score. As a second performance measure, we compute metrics from the Receiver Operating Characteristic curve (ROC). To this end, we integrate over the area under the ROC curve to get the AUROC metric and identify the False Postive Rate (FPR) at $95\%$ True Postive Rate (TPR), namely FPR-95\%-TPR.

\subsection{Implementation Details}

We follow~\cite{liang2020learning} and transform all agents in a local coordinate system, where the target agent is centred at $t=0$. The positive x-axis is defined as the orientation of the target agent, estimated from the location at $t=-1$ to the location at $t=0$. The Seq2Seq and STGAE models use absolute coordinates in this local coordinate system, while for the LaneGCN-AE we compute positional displacements, which are proportional to the velocity, by subtracting locations of consecutive time steps. We only consider actors which are present in the last frame of the sequence. For the agents which appear in the middle of a sequence, we use zero padding to fill the missing states in the beginning of the sequence. Disappearing agents are ignored, since they have less influence and often disappear in far distance from the target agent. To include map information in LaneGCN-AE, we compute the lane graph as proposed in~\cite{liang2020learning}.

CVM and LTI approximate the trajectory with linear equations. The CVM extrapolates the velocity estimated from the locations at the first two time steps $t=-T+1$ and $t=-T+2$ over sequence length $T$. The LTI model computes equidistantly sampled locations on a straight line between the start location at $t=-T+1$ and the end location at $t=0$. For the Seq2Seq auto-encoder we employ the symmetric encoder-decoder architecture as proposed in~\cite{chang2019argoverse} with $8$ linear embedding and $16$ hidden LSTM units. We use the same Seq2Seq auto-encoder for the STGAE, but with the interaction-aware features from the spatial graph embedding as input. The spatial graph embedding layer transforms the trajectories of all agents into an embedding space and aggregates neighboring features using the distance-based adjacency matrix.
The LaneGCN-AE follows the structure of the original LaneGCN~\cite{liang2020learning}. We remove the multi-modal prediction and employ a single head for trajectory reconstruction given the actor feature vector from the FusionNet. We decrease the actor feature space from $128$ to $16$ to limit the representation space of the auto-encoder. We train all deep auto-encoders with the Adam optimiser~\cite{kingma2014adam} over $36$ epochs and a batch size of $32$. After $32$ epochs the learning rate gets updated from $10^{-3}$ to $10^{-4}$. For subsequent evaluation, we always pick the model of the epoch with the lowest loss on the validation set. After training the auto-encoders we train the OC-SVM variants. The OC-SVM is implemented with a Gaussian kernel and hyperparameters are selected using grid search with $\gamma \in \{2^{-10},2^{-9},...,2^{-1}\}$ and $\nu \in \{0.01, 0.1\}$. For validation of the hyperparamters, we randomly select \SI{20}{\%} of the annotated test data. Therefore OC-SVM has a minor supervised advantage. The latent space centre $\mathbf{c}$ for the DSVDD loss is initialised as the mean of the latent representations of all training samples computed by an initial forward pass~\cite{pmlr-v80-ruff18a}.

\begin{table*}[t]
\caption{Comparison of the baselines using the defined anomaly detection metrics. 
Best values highlighted in bold.}
\centering
\resizebox{\textwidth}{!}{\begin{tabular}{p{0.06\linewidth}ccclcccc}
\toprule
 & \textbf{Category}            & \textbf{\begin{tabular}[c]{@{}c@{}}Using \\ dyn. Context\end{tabular}} &\textbf{\begin{tabular}[c]{@{}c@{}}Using \\ stat. Context\end{tabular}} & \textbf{Method}       & \textbf{AUPR-Abnormal} $\uparrow$ & \textbf{AUPR-Normal} $\uparrow$ & \textbf{AUROC} $\uparrow$ & \textbf{FPR-95\%-TPR} $\downarrow$ \\ 
 \midrule
 \parbox[t]{5mm}{\multirow{5}{*}{\rotatebox[origin=c]{90}{\begin{tabular}[c]{@{}c@{}}{Recon-}\\{struction}\end{tabular}}}}
 & \multirow{2}{*}{Linear}   & \xmark  & \xmark                                                                  & CVM~\cite{scholler2020constant}                   &      47.19                  & 86.00                      & 72.30                & 81.20                       \\
 &                           & \xmark  & \xmark                                                                 & LTI                   & 50.45                        & 85.71                      & 73.14                & 82.22                 
 \\ 
\cmidrule(lr{1em}){2-9}
 & \multirow{3}{*}{Deep}     & \xmark  & \xmark                                                                 & Seq2Seq~\cite{chang2019argoverse}
 &  59.21                      & \textbf{88.07}                      & 76.56                & 77.62                       \\

&                         &  \cmark   & \xmark                                                       
   & STGAE~\cite{pmlr-v164-wiederer22a}
 & \textbf{59.65}                       & 87.85                      & \textbf{76.75}                & 76.48 \\

 &                           &  \cmark & \cmark                                                                 & LaneGCN-AE~\cite{liang2020learning}
 & 57.19                        & 87.22                      & 75.25                & \textbf{75.94}                        
\\ \midrule
 \parbox[t]{5mm}{\multirow{6}{*}{\rotatebox[origin=c]{90}{\begin{tabular}[c]{@{}c@{}}{One-}\\{Class}\end{tabular}}}}
 & \multirow{3}{*}{Two-stage}  & \xmark & \xmark                                                                 & Seq2Seq+OC-SVM
 & 34.47                        & 70.25                      & 50.47                & 98.33                       \\
  &                            & \cmark & \xmark                                                                 & STGAE+OC-SVM~\cite{pmlr-v164-wiederer22a}
& 33.32                       & 77.71                      & 59.16                & 91.27                       \\
 &                            & \cmark & \cmark                                                                 & LaneGCN-AE+OC-SVM
 & 51.88                       & 86.93                      & 72.94                & 82.02                       \\ 
\cmidrule(lr{1em}){2-9}
 & \multirow{3}{*}{End-to-End} 
& \cmark & \xmark                                                                 & Seq2Seq+DSVDD     
& 51.37                       & 82.47                      & 69.34                & 88.79                       \\ 

   &                            & \cmark & \xmark                                                   
   & STGAE+DSVDD       
& 48.09                        & 83.59                      & 69.65                & 85.44                       \\ 
 &                             & \cmark & \cmark                                                                 & LaneGCN-AE+DSVDD       
 & 53.14                       &  85.21                    & 72.33                & 85.55                       \\ 
\bottomrule
\end{tabular}}
\label{tab:results}
\end{table*}

\begin{figure*}[t]
\centering
    \subfloat[Abnormal scenario: Cancel turn.]{
        \includegraphics[width=0.32\textwidth]{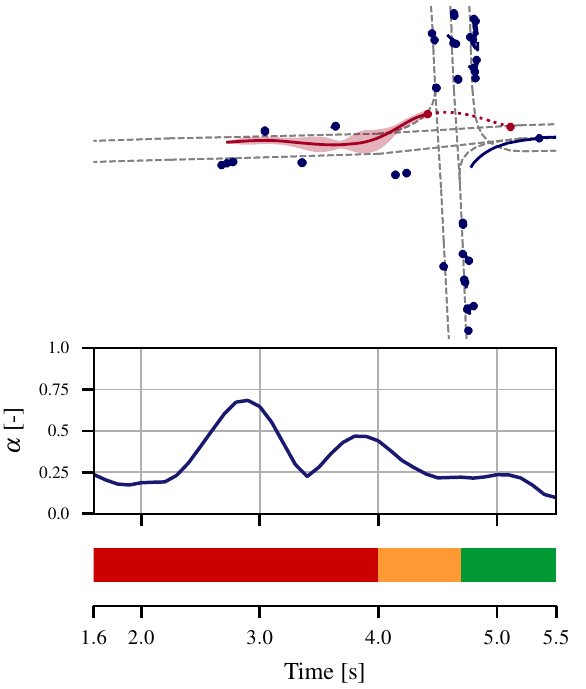}
        \label{fig:res_abnormal_1}
    }
    \subfloat[Abnormal scenario: Last minute turn.]{
        \includegraphics[width=0.32\textwidth]{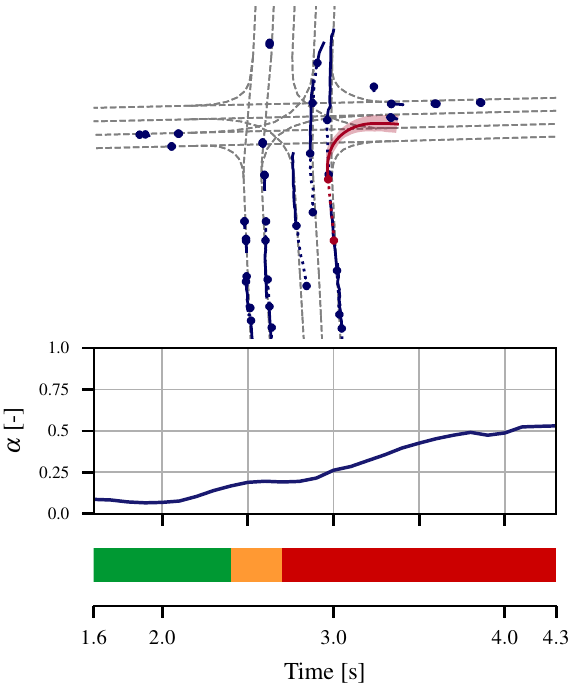}
        \label{fig:res_abnormal_2}
    }
    \subfloat[Normal scenario: Lane change right.]{
        \includegraphics[width=0.32\textwidth]{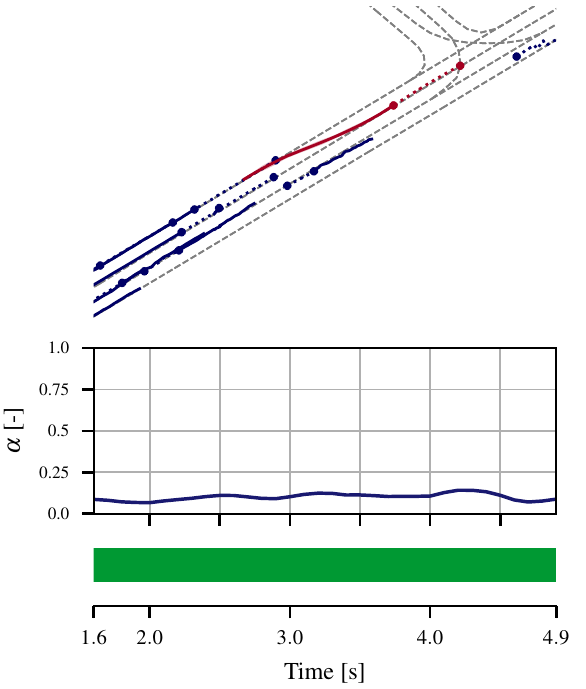}
        \label{fig:res_normal_1}
    }
    \caption{Qualitative anomaly detection results of the STGAE model. At the top we show the scene visualised by the street centre lines as well as the actor trajectories, with the target agent in red and all other actors in blue. Trajectories of the first $1.6$ seconds of the sequence are dashed. The model predicts an anomaly score $\alpha$ for the remaining time, which is illustrated as error band around the trajectory of the target agent. In the middle and the bottom, the anomaly score $\alpha$, normalised by the highest score in the test set, and the ground truth annotation are visualised over time, respectively. For the ground truth we encode normal behaviour labels in green, abnormal behaviour labels in red and ignore regions in orange. The anomaly score increases for the abnormal behaviour in Figure~\ref{fig:res_abnormal_1} and~\ref{fig:res_abnormal_2} and remains low for the normal lane change manoeuvre in Figure~\ref{fig:res_normal_1}.}
    \label{fig:qual_results}
\end{figure*}

\section{Results}

We compare the results of all baselines on our presented benchmark. In addition, we illustrate qualitative results for different abnormal and normal situations. 

\textbf{Comparison of Baselines.} Table~\ref{tab:results} shows the results of all baselines, cross and check marks indicate the use of dynamic or static context information in addition to the target agent trajectory. In total we compare the results of $11$ methods including the reconstruction and the one-class classification methods. Of all model types the deep auto-encoder networks show the best results. In particular the~\textit{STGAE} wins in two out of four metrics including our main metric AUPR-abnormal where it reaches \SI{59.65}{\%}. \textit{Seq2Seq} and ~\textit{LaneGCN-AE} are on par with their performances. Interestingly, both interaction-aware models, \textit{STGAE} and \textit{LaneGCN-AE}, cannot improve the results further using additional dynamic and static context information in comparison to the single-agent \textit{Seq2Seq} model. Algorithms, which learn useful features from the additional context, e.g. by employing new loss functions, are required to improve the anomaly detection quality. We leave this as a future research direction. In general, the linear models fall behind the deep reconstruction methods.
The assumption that anomalies in multi-agent trajectories are linearly separable from the normal driving is too simple, resulting in low performance for the linear models. It is only correct for a few anomalies with high dynamics, i.e. anomalies characterised by high acceleration or sharp turns, but does not generalise to complex driving anomalies.

Compared to each other, the one-class methods do not show consistent results. In particular the OC-SVM variants show lower values in the metrics compared to the deep reconstruction methods. The OC-SVM is not able to learn a useful representation of the normal data neither using the latent features from the \textit{Seq2Seq} nor from \textit{STGAE} model. The separation of anomalies seems easier in the output space, as for the deep reconstruction methods, than in the latent space, as for the OC-SVM-based methods. 
Other for the DSVDD variants, after training jointly on the reconstruction and the DSVDD loss, the similarity between a latent feature vector and the hypersphere centre is a reasonable measure to detect anomalies. Still, they cannot reach the performance of the deep reconstruction methods. 
Especially the deep reconstruction methods and the end-to-end one-class methods are valuable baselines for multi-agent anomaly detection.

\textbf{Qualitative Results.} Figure~\ref{fig:qual_results} shows qualitative results of the \textit{STGAE} model for two abnormal and one normal scene. We observe the increasing anomaly score in the abnormal scenarios. In Figure~\ref{fig:res_abnormal_1} the target vehicle cancels a turn in the middle of the intersection. The dynamics in the trajectory highly deviates from the normal driving in the training data. Therefore, the \textit{STGAE} is not able to fully reconstruct the input, which leads to an increase of the reconstruction error and consequently of the anomaly score. Similar, the anomaly score in Figure~\ref{fig:res_abnormal_2} increases. It shows a last minute turn at the end of the sequence. This is not the case for the normal scenario. The manoeuvre of a lane change is reconstructed with low reconstruction error by the \textit{STGAE} model resulting in a low anomaly score during the entire scene.

\section{Conclusion}
We present a benchmark for anomaly detection in multi-agent driving trajectories. Our main contribution is the R-U-MAAD benchmark that builds on top of the Argoverse dataset with multiple interactive and non-interactive anomalies in diverse road scenarios. We provide 11 anomaly detection baseline models including linear methods, deep auto-encoders as well as deep one-class classification models. The deep baselines learn a representation of normal driving behaviour from the Argoverse training data in an unsupervised manner. Using this representation, they recognise anomalies in novel scenes during testing. With the presented work we hope to advance research in the area of multi-agent anomaly detection and want to establish a benchmark for the comparison of future algorithms.

\addtolength{\textheight}{-1.0cm} %

\section*{Acknowledgment}
The research leading to these results is funded by the German Federal Ministry for Economic Affairs and Energy within the project “KI Delta Learning" (Förderkennzeichen 19A19013A). The authors would like to thank the consortium for the successful cooperation.

\bibliographystyle{IEEEtranS}
\bibliography{references}  %

\begin{thebibliography}{10}
\providecommand{\url}[1]{#1}
\csname url@rmstyle\endcsname
\providecommand{\newblock}{\relax}
\providecommand{\bibinfo}[2]{#2}
\providecommand\BIBentrySTDinterwordspacing{\spaceskip=0pt\relax}
\providecommand\BIBentryALTinterwordstretchfactor{4}
\providecommand\BIBentryALTinterwordspacing{\spaceskip=\fontdimen2\font plus
\BIBentryALTinterwordstretchfactor\fontdimen3\font minus
  \fontdimen4\font\relax}
\providecommand\BIBforeignlanguage[2]{{%
\expandafter\ifx\csname l@#1\endcsname\relax
\typeout{** WARNING: IEEEtran.bst: No hyphenation pattern has been}%
\typeout{** loaded for the language `#1'. Using the pattern for}%
\typeout{** the default language instead.}%
\else
\language=\csname l@#1\endcsname
\fi
#2}}

\bibitem{branco2020interleaved}
B.~Branco, P.~Abreu, A.~S. Gomes, M.~S. Almeida, J.~T. Ascens{\~a}o, and
  P.~Bizarro, ``Interleaved sequence rnns for fraud detection,'' in
  \emph{Proceedings of the 26th ACM SIGKDD international conference on
  knowledge discovery \& data mining}, 2020, pp. 3101--3109.

\bibitem{brockman2016openai}
G.~Brockman, V.~Cheung, L.~Pettersson, J.~Schneider, J.~Schulman, J.~Tang, and
  W.~Zaremba, ``Openai gym,'' \emph{arXiv preprint arXiv:1606.01540}, 2016.

\bibitem{brugman2004annotating}
H.~Brugman, A.~Russel, and X.~Nijmegen, ``Annotating multi-media/multi-modal
  resources with elan.'' in \emph{LREC}, 2004, pp. 2065--2068.

\bibitem{nuscenes}
H.~Caesar, V.~Bankiti, A.~H. Lang, S.~Vora, V.~E. Liong, Q.~Xu, A.~Krishnan,
  Y.~Pan, G.~Baldan, and O.~Beijbom, ``nuscenes: A multimodal dataset for
  autonomous driving,'' in \emph{CVPR}, 2020.

\bibitem{chang2019argoverse}
M.-F. Chang, J.~Lambert, P.~Sangkloy, J.~Singh, S.~Bak, A.~Hartnett, D.~Wang,
  P.~Carr, S.~Lucey, D.~Ramanan, \emph{et~al.}, ``Argoverse: 3d tracking and
  forecasting with rich maps,'' in \emph{CVPR}, 2019.

\bibitem{cheifetz2011apattern}
N.~Cheifetz, A.~Samé, P.~Aknin, and E.~de~Verdalle, ``A pattern recognition
  approach for anomaly detection on buses brake system,'' in \emph{ITSC}, 2011,
  pp. 266--271.

\bibitem{corbiere2019addressing}
C.~Corbi{\`e}re, N.~Thome, A.~Bar-Hen, M.~Cord, and P.~P{\'e}rez, ``Addressing
  failure prediction by learning model confidence,'' \emph{Advances in Neural
  Information Processing Systems}, vol.~32, 2019.

\bibitem{eskin2002geometric}
E.~Eskin, A.~Arnold, M.~Prerau, L.~Portnoy, and S.~Stolfo, ``A geometric
  framework for unsupervised anomaly detection,'' in \emph{Applications of data
  mining in computer security}.\hskip 1em plus 0.5em minus 0.4em\relax
  Springer, 2002, pp. 77--101.

\bibitem{ettinger2021large}
S.~Ettinger, S.~Cheng, B.~Caine, C.~Liu, H.~Zhao, S.~Pradhan, Y.~Chai, B.~Sapp,
  C.~R. Qi, Y.~Zhou, \emph{et~al.}, ``Large scale interactive motion
  forecasting for autonomous driving: The waymo open motion dataset,'' in
  \emph{CVPR}, 2021, pp. 9710--9719.

\bibitem{gao2020vectornet}
J.~Gao, C.~Sun, H.~Zhao, Y.~Shen, D.~Anguelov, C.~Li, and C.~Schmid,
  ``Vectornet: Encoding hd maps and agent dynamics from vectorized
  representation,'' in \emph{Proceedings of the IEEE/CVF Conference on Computer
  Vision and Pattern Recognition}, 2020, pp. 11\,525--11\,533.

\bibitem{houston2020lyft}
J.~Houston, G.~Zuidhof, L.~Bergamini, Y.~Ye, A.~Jain, S.~Omari, V.~Iglovikov,
  and P.~Ondruska, ``One thousand and one hours: Self-driving motion prediction
  dataset,'' 2020.

\bibitem{kingma2014adam}
D.~P. Kingma and J.~Ba, ``Adam: A method for stochastic optimization,''
  \emph{arXiv preprint arXiv:1412.6980}, 2014.

\bibitem{li2021unsupervised}
D.~Li, H.~Guo, Z.~Wang, and Z.~Zheng, ``Unsupervised fake news detection based
  on autoencoder,'' \emph{IEEE Access}, vol.~9, pp. 29\,356--29\,365, 2021.

\bibitem{liang2020learning}
M.~Liang, B.~Yang, R.~Hu, Y.~Chen, R.~Liao, S.~Feng, and R.~Urtasun, ``Learning
  lane graph representations for motion forecasting,'' in \emph{ECCV}, 2020.

\bibitem{ma2019trafficpredict}
Y.~Ma, X.~Zhu, S.~Zhang, R.~Yang, W.~Wang, and D.~Manocha, ``Trafficpredict:
  Trajectory prediction for heterogeneous traffic-agents,'' in
  \emph{Proceedings of the AAAI Conference on Artificial Intelligence},
  vol.~33, 2019, pp. 6120--6127.

\bibitem{makansi2022you}
\BIBentryALTinterwordspacing
O.~Makansi, J.~V. K{\"u}gelgen, F.~Locatello, P.~V. Gehler, D.~Janzing,
  T.~Brox, and B.~Sch{\"o}lkopf, ``You mostly walk alone: Analyzing feature
  attribution in trajectory prediction,'' in \emph{International Conference on
  Learning Representations}, 2022. [Online]. Available:
  \url{https://openreview.net/forum?id=POxF-LEqnF}
\BIBentrySTDinterwordspacing

\bibitem{matousek2019detecting}
M.~Matousek, M.~EL-Zohairy, A.~Al-Momani, F.~Kargl, and C.~Bösch, ``Detecting
  anomalous driving behavior using neural networks,'' in \emph{2019 IEEE
  Intelligent Vehicles Symposium (IV)}, 2019, pp. 2229--2235.

\bibitem{oikawa2020fast}
H.~Oikawa, T.~Nishida, R.~Sakamoto, H.~Matsutani, and M.~Kondo, ``Fast
  semi-supervised anomaly detection of drivers’ behavior using online
  sequential extreme learning machine,'' in \emph{ITSC}, 2020, pp. 1--8.

\bibitem{pmlr-v80-ruff18a}
L.~Ruff, R.~Vandermeulen, N.~Goernitz, L.~Deecke, S.~A. Siddiqui, A.~Binder,
  E.~M{\"u}ller, and M.~Kloft, ``Deep one-class classification,'' in
  \emph{Proceedings of the 35th International Conference on Machine Learning},
  ser. Proceedings of Machine Learning Research, J.~Dy and A.~Krause, Eds.,
  vol.~80.\hskip 1em plus 0.5em minus 0.4em\relax PMLR, 10--15 Jul 2018, pp.
  4393--4402.

\bibitem{sabour2021deepflow}
S.~Sabour, S.~Rao, and M.~Ghaderi, ``Deepflow: Abnormal traffic flow detection
  using siamese networks,'' in \emph{2021 IEEE International Smart Cities
  Conference (ISC2)}.\hskip 1em plus 0.5em minus 0.4em\relax IEEE, 2021, pp.
  1--7.

\bibitem{schmidt2022cratpred}
J.~Schmidt, J.~Jordan, F.~Gritschneder, and K.~Dietmayer, ``Crat-pred: Vehicle
  trajectory prediction with crystal graph convolutional neural networks and
  multi-head self-attention,'' in \emph{ICRA}, 2022, pp. 7799--7805.

\bibitem{scholkopf1999support}
B.~Sch{\"o}lkopf, R.~C. Williamson, A.~J. Smola, J.~Shawe-Taylor, J.~C. Platt,
  \emph{et~al.}, ``Support vector method for novelty detection.'' in
  \emph{NIPS}, vol.~12.\hskip 1em plus 0.5em minus 0.4em\relax Citeseer, 1999,
  pp. 582--588.

\bibitem{scholler2020constant}
C.~Sch{\"o}ller, V.~Aravantinos, F.~Lay, and A.~Knoll, ``What the constant
  velocity model can teach us about pedestrian motion prediction,'' \emph{IEEE
  Robotics and Automation Letters}, vol.~5, no.~2, pp. 1696--1703, 2020.

\bibitem{schreiber2020motion}
M.~Schreiber, V.~Belagiannis, C.~Gläser, and K.~Dietmayer, ``Motion estimation
  in occupancy grid maps in stationary settings using recurrent neural
  networks,'' in \emph{2020 IEEE International Conference on Robotics and
  Automation (ICRA)}, 2020, pp. 8587--8593.

\bibitem{shah2021asimulation}
T.~Shah, J.~R. Lepird, A.~T. Hartnett, and J.~M. Dolan, ``A simulation-based
  benchmark for behavioral anomaly detection in autonomous vehicles,'' in
  \emph{ITSC}, 2021, pp. 2074--2081.

\bibitem{strohbeck2020Multiple}
J.~Strohbeck, V.~Belagiannis, J.~Müller, M.~Schreiber, M.~Herrmann, D.~Wolf,
  and M.~Buchholz, ``Multiple trajectory prediction with deep temporal and
  spatial convolutional neural networks,'' in \emph{2020 IEEE/RSJ International
  Conference on Intelligent Robots and Systems (IROS)}, 2020, pp. 1992--1998.

\bibitem{tian2021constrained}
Y.~Tian, G.~Pang, F.~Liu, Y.~Chen, S.~H. Shin, J.~W. Verjans, R.~Singh, and
  G.~Carneiro, ``Constrained contrastive distribution learning for unsupervised
  anomaly detection and localisation in medical images,'' in
  \emph{International Conference on Medical Image Computing and
  Computer-Assisted Intervention}.\hskip 1em plus 0.5em minus 0.4em\relax
  Springer, 2021, pp. 128--140.

\bibitem{ucar2021differential}
S.~Ucar, B.~Hoh, and K.~Oguchi, ``Differential deviation based abnormal driving
  behavior detection,'' in \emph{ITSC}, 2021, pp. 1553--1558.

\bibitem{wang2016statistical}
W.~Wang, J.~Xi, and X.~Li, ``Statistical pattern recognition for driving styles
  based on bayesian probability and kernel density estimation,'' \emph{arXiv
  preprint arXiv:1606.01284}, 2016.

\bibitem{weis2017anomaly}
T.~Weis, M.~Mundt, P.~Harding, and V.~Ramesh, ``Anomaly detection for
  automotive visual signal transition estimation,'' in \emph{2017 IEEE 20th
  International Conference on Intelligent Transportation Systems (ITSC)}, 2017,
  pp. 1--8.

\bibitem{wiederer2020traffic}
J.~Wiederer, A.~Bouazizi, U.~Kressel, and V.~Belagiannis, ``Traffic control
  gesture recognition for autonomous vehicles,'' in \emph{2020 IEEE/RSJ
  International Conference on Intelligent Robots and Systems (IROS)}.\hskip 1em
  plus 0.5em minus 0.4em\relax IEEE, 2020, pp. 10\,676--10\,683.

\bibitem{pmlr-v164-wiederer22a}
J.~Wiederer, A.~Bouazizi, M.~Troina, U.~Kressel, and V.~Belagiannis, ``Anomaly
  detection in multi-agent trajectories for automated driving,'' in
  \emph{Proceedings of the 5th Conference on Robot Learning}, 2022, pp.
  1223--1233.

\bibitem{yu2020bdd100k}
F.~Yu, H.~Chen, X.~Wang, W.~Xian, Y.~Chen, F.~Liu, V.~Madhavan, and T.~Darrell,
  ``Bdd100k: A diverse driving dataset for heterogeneous multitask learning,''
  in \emph{Proceedings of the IEEE/CVF conference on computer vision and
  pattern recognition}, 2020, pp. 2636--2645.

\bibitem{zaheer2020old}
M.~Z. Zaheer, J.-h. Lee, M.~Astrid, and S.-I. Lee, ``Old is gold: Redefining
  the adversarially learned one-class classifier training paradigm,'' in
  \emph{CVPR}, 2020, pp. 14\,183--14\,193.

\bibitem{interactiondataset}
W.~Zhan, L.~Sun, D.~Wang, H.~Shi, A.~Clausse, M.~Naumann, J.~K\"ummerle,
  H.~K\"onigshof, C.~Stiller, A.~de~La~Fortelle, and M.~Tomizuka,
  ``{INTERACTION} {Dataset}: {An} {INTERnational}, {Adversarial} and
  {Cooperative} {moTION} {Dataset} in {Interactive} {Driving} {Scenarios} with
  {Semantic} {Maps},'' \emph{arXiv:1910.03088 [cs, eess]}, Sept. 2019.

\bibitem{zhu2018adeep}
L.~Zhu, F.~Guo, R.~Krishnan, and J.~W. Polak, ``A deep learning approach for
  traffic incident detection in urban networks,'' in \emph{2018 21st
  International Conference on Intelligent Transportation Systems (ITSC)}, 2018,
  pp. 1011--1016.

\bibitem{zong2018deep}
B.~Zong, Q.~Song, M.~R. Min, W.~Cheng, C.~Lumezanu, D.~Cho, and H.~Chen, ``Deep
  autoencoding gaussian mixture model for unsupervised anomaly detection,'' in
  \emph{International conference on learning representations}, 2018.

\end{thebibliography}

\end{document}